\documentclass[10pt,twocolumn,letterpaper]{article}

\usepackage{iccv}
\usepackage{subfigure}
\usepackage{booktabs} 
\usepackage{hyperref}
\usepackage{url}
\usepackage{comment}
\usepackage{amsmath,amssymb} 
\usepackage{color}
\usepackage{wrapfig}
\usepackage{hyperref}

\usepackage{times}
\usepackage{graphicx}

\iccvfinalcopy 

\def\iccvPaperID{1964} 

\ificcvfinal\fi

\begin{document}

\title{Multilayer Dense Connections for Hierarchical Concept Prediction}

\author{Toufiq Parag \qquad Hongcheng Wang\\
Comcast Applied AI Research\\
Washington D.C., USA.\\
{\tt\small toufiq.parag@gmail.com}
}

\maketitle
\ificcvfinal\thispagestyle{empty}\fi

\begin{abstract}
Multinomial logistic regression with a single final layer of dense connections has become the ubiquitous technique for CNN-based classification. While these classifiers project a mapping between the input and a set of output category classes, they do not typically yield  a comprehensive description of the category. In particular, when a CNN based image classifier correctly identifies the image of a Chimpanzee, its output does not clarify that Chimpanzee is a member of Primate, Mammal, Chordate families and a living thing. We propose a multilayer dense connectivity for concurrent prediction of category \emph{and} its conceptual superclasses in hierarchical order by the same CNN. We experimentally demonstrate that our proposed network can simultaneously predict both the coarse superclasses and finer categories better than several existing algorithms in multiple datasets.
\end{abstract}

\section{Introduction}

Classification is a core concept for numerous computer vision tasks. Given the convolutional features, different architectures classify either the image itself~\cite{resnet, inceptionv4}, the region/bounding boxes for object detection~\cite{mask-rcnn, ssd}, or, at the granular level, pixels for scene segmentation~\cite{liang18deeplabv3}.  Although early image recognition works employed multilayer classification layers~\cite{alexnet, vgg}, the more recent models have all been using single layer dense connection~\cite{resnet2, inceptionv4} or convolutions~\cite{lin2017focal}.


The vision community has invented a multitude of techniques to enhance the capacity of feature computation layers~\cite{resnetxt,densenet,squeezenet,deformable,xception,efficientnet}. But, the classification layer has mostly retained the form  of a multinomial/softmax logistic regression performing a mapping  from a set of input images to a set of categories. Despite achieving impressive accuracy in classifying target categories, these networks do not furnish a comprehensive depiction of the input entity through their outputs. In particular, when a CNN correctly identifies an image of an English Setter, it is not laid out in the output that it is an instance of a dog, or more extensively, a hunting dog, a domestic animal and a living thing. It is rational to assume that convolutional layers construct some internal representation of the conceptual (coarser) superclasses, e.g., dog, animal etc., during training. Not attempting to classify the coarse superclasses along with the finer categories amounts to a waste of information already computed within the convolutional layers.
\begin{figure}
\vspace{-0.05in}
\begin{center}
\includegraphics[width=0.32\columnwidth,height=0.37\columnwidth]{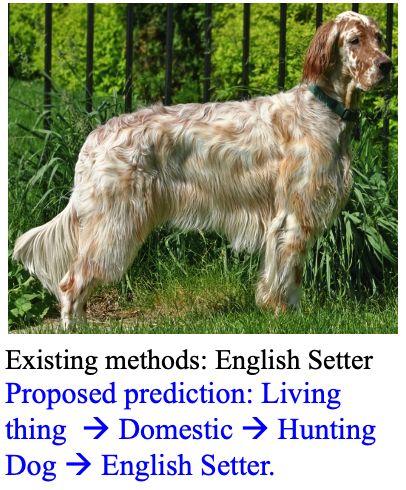}~
\includegraphics[width=0.32\columnwidth,height=0.37\columnwidth]{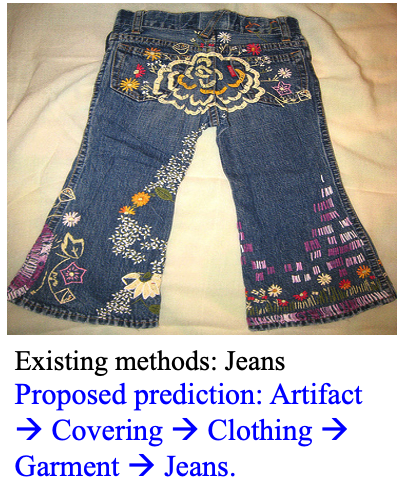}~
\includegraphics[width=0.32\columnwidth,height=0.37\columnwidth]{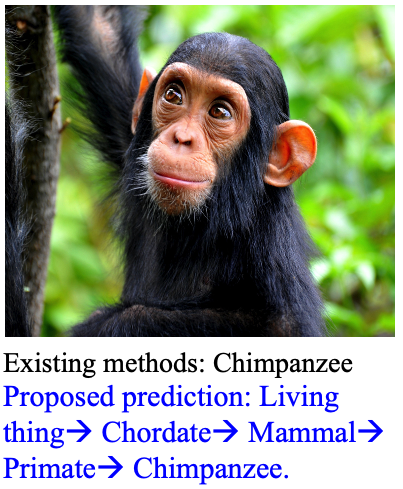}
\end{center}
\vspace{-0.3cm}
\label{F:INTRO}
\caption{\small The goal of the proposed algorithm. In contrast to the existing methods, our proposed CNN architecture predicts the chain of superclass concepts  as well as  the finer category.}
\vspace{-0.5cm}
\end{figure}

In fact, there are practical applications for producing the conceptual superclasses as well as the finer categories in the output. \textbf{(1)} Furnishing coarser labels improves interpretability of the classifier performance. As Bertinetto et al.~\cite{bertinetto20cvpr} also imply, even if an eagle is misclassified as a parrot, the capability of inferring that it is a bird, and not an artifact (e.g.,  drone), may be beneficial in some applications (e.g., surveillance). \textbf{(2)} An object detector can enhance its capability on unseen categories by adopting the proposed classification scheme. For example, a movie/TV violence detection tool can classify a lightsaber to coarse 'weapon' superclass although it is perhaps too rare to include in the training set as a category . \textbf{(3)} In visual question answering (VQA),  encoding concept classes could expand the scope of query terms by allowing broader description ('how many vehicles are present' in addition to 'how many buses', 'how many trucks' etc.)~\cite{cao18vqr,fvqa}. 


Simultaneous classification of coarse and finer classes requires a relationship among the conceptual superclasses and categories. Ontologies of WordNet~\cite{wordnet} or biological taxonomy~\cite{inaturalist} encode the hierarchical organization of finer category classes (e.g., English Setter) and their conceptual superclasses (e.g., Hunting dog, Domestic animal, Living thing).  Several past studies proposed theoretically/conceptually elegant methods to exploit concept hierarchies~\cite{deng14hex, ding15phex, yan15hdcnn, bertinetto20cvpr}. However, the goal of these methods is to utilize the ontology  to improve the categorical classification  -- not to predict the coarse concepts. 

There exists sophisticated techniques that directly predict, or may be adapted to predict, the superclasses~\cite{hu16cvpr, yan15hdcnn}. Notable among them, Yan et al.~\cite{yan15hdcnn} employ a shallow single layer hierarchy but does not illustrate how to expand it to multilevel ontology. On the other hand, the model of Hu et al.~\cite{hu16cvpr} is perhaps more complex than is necessary for the task. Many of these models employ a separate tool -- e.g., RNN or RNN-like network in~\cite{Guo2017CNNRNNAL, hu16cvpr} and CRFs in~\cite{deng14hex, ding15phex} -- that imposes additional difficulty in training/tuning.  We have not found an existing (deep learning) model that attempts to predict both the finer categories and the chain of (multiple) ancestor concepts for an input image by a single CNN.  The classical (pre-CNN) hedging method~\cite{deng12hedge} computes either the finer labels or one of its superclasses exclusively, but not both simultaneously.

In this paper, we introduce a CNN to classify the category \emph{and} the concept superclasses simultaneously. As illustrated in Figure~\ref{F:INTRO}, in order to classify any category class (e.g., English Setter), our model is constrained to also predict the ancestor superclasses (e.g., Hunting dog, Domestic animal, Living thing) in the same order as defined in a given ontology(e.g., WordNet or biological taxonomy). We propose a configuration of multilayer dense connections to predict the category \& concept superclasses as well as model their interrelations based on the ontology. We also propose a simple method to prune and rearrange the label hierarchy for efficient connectivity.

Capturing the hierarchical relationship within the CNN architecture itself enables us to train the model end-to-end (as opposed to attaching a separate tool). Our multilayer connections can be easily attached to any existing CNNs (e.g., ResNet-50, Inception V4) by replacing the single layer multinomial regression and we experimentally demonstrate such a network can be learned by standard training protocols on multiple datasets without sacrificing  category-wise accuracy. More importantly, our experiments validate the necessity of the proposed multilayer model for concept prediction in comparison to the flat single layer connections (Sections~\ref{S:IMGNET},~\ref{S:IMGNET-ADV},~\ref{S:AWA}). 

Furthermore, we also show that our model outperforms an improved version of~\cite{deng12hedge} adapted for concurrent concept and category prediction both on ImageNet dataset(Section~\ref{S:IMGNET}) and the challenging images of~\cite{hendrycks19adv} (Section~\ref{S:IMGNET-ADV}). Experiments on the AwA2 dataset (Section~\ref{S:AWA}) illustrate that the proposed network yields accuracies superior to those of~\cite{hu16cvpr,yan15hdcnn} while being lighter than both of them. Finally, we demonstrate that our model inherently makes better mistakes than a CNN trained with hierarchical cross entropy~\cite{bertinetto20cvpr} on the iNaturalist dataset (Section~\ref{S:INAT}).

\section{Relevant works}\label{S:RELEVANT_WORKS}
Use of hierarchical classifiers can be traced back to the early works of~\cite{torralba04boosting,wu04cascade,fergus2010SemanticLS} that shared features for improved classification. Some studies drew inspiration for using a hierarchical organization of categories from human cognitive systems~\cite{zhao11largescale, deng10K} while others suggested practical benefits of label hierarchies with with tens of thousands of categories~\cite{bengio10nips,deng11labeltree}.

The classical algorithm of \cite{deng12hedge} aim to  predict either a coarse concept or a fine category exclusively by maximizing the reward based on aggregated probabilities in a label hierarchy. The reported results suggest the prediction of superclasses comes at the expense of the fine level category classification failure. For CNN based classification, Deng et al.~\cite{deng14hex} modeled the relationships such as subsumption, overlap and exclusion among the categories via a CRF. Although the CRF parameter can be trained via gradient descent, the inference required a separate computation of message passing. The work of~\cite{ding15phex} extended this model by utilizing probabilistic label relationships. 


The HDCNN framework~\cite{yan15hdcnn} creates its own single layer hierarchy by grouping the finer classes into coarse concepts. The framework comprises two modules for coarse and fine classes where the coarse prediction modulates the layers for finer classification. Hu et al.~\cite{hu16cvpr} present a network structure inspired by the bidirectional recurrent networks to model the hierarchical interactions. The model leads to a large number of inter and intra-layer label interactions some of which needed to be manually hard-coded to 0. Guo et al.\cite{Guo2017CNNRNNAL} proposed to apply an RNN for conceptual superclass prediction, however, it does not clarify how the hierarchy was generated and does not report performance on large dataset (e.g. ImageNet).


\begin{figure*}[t]
\vspace{-0.2in}
\begin{center}
\includegraphics[width=0.45\textwidth,height=0.31\textwidth]{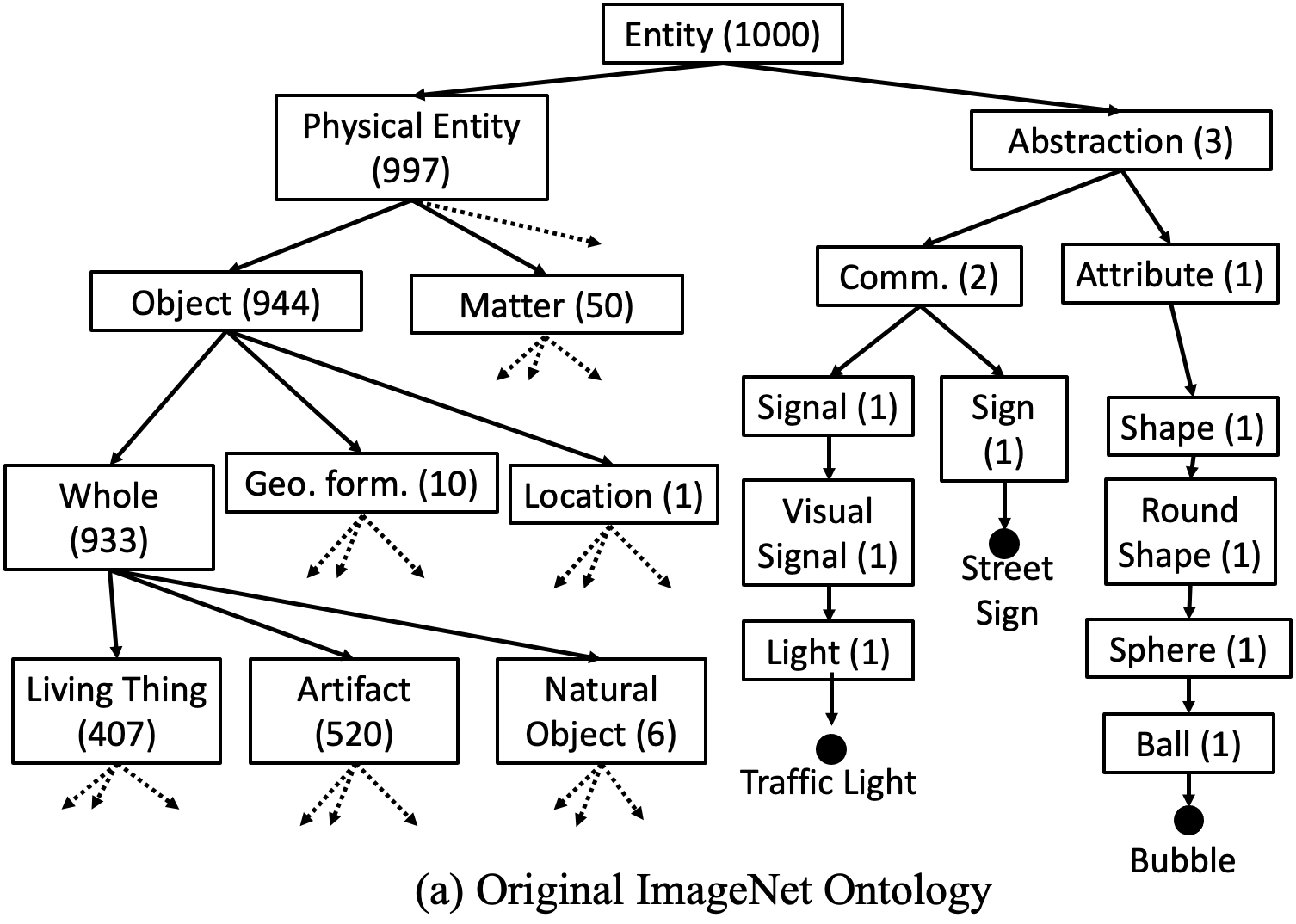}
\qquad \includegraphics[width=0.45\textwidth,height=0.3\textwidth]{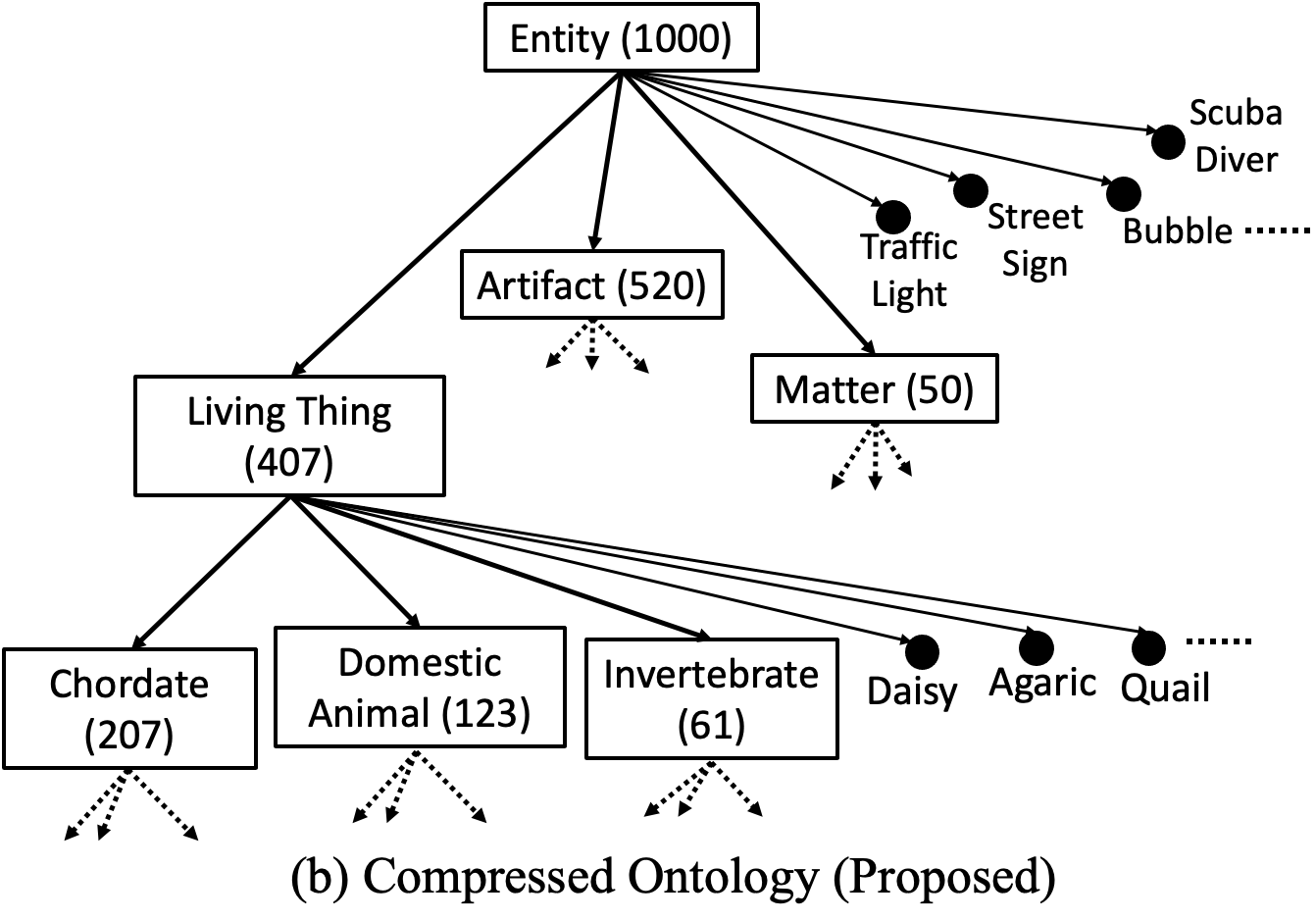}
\end{center}
\vspace{-0.4cm}
\caption{\small  Partial view of the original (left) and condensed (right) label hierarchies. Concepts are enclosed in rectangular boxes, with number of all descendants in parentheses. All the parent concepts of the categories Traffic Light, Street Sign and Bubble in the original ontology are absorbed to Entity in the compressed ontology by removing the redundant single parent-child connections and excluding nodes with descendant count $< \delta$. Similarly, the ancestors (Physical Entity, Object, Whole) of Living Thing, Artifact are all collapsed to Entity because $\tau \%$ descendants of each parent in the ancestry were also descendants of its child.}
\label{F:HIER}
\vspace{-0.4cm}
\end{figure*}

Many past works target improving the fine category classification by using the concept hierarchy as auxiliary source of information~\cite{deng14hex,ding15phex, yan15hdcnn, bertinetto20cvpr}. These studies do not explicitly provide a mechanism to predict both coarse superclasses and finer categories as we propose. Some existing methods can be modified to yield ancestor concepts as output~\cite{deng12hedge, hu16cvpr, yan15hdcnn} -- we have compared the performances of these models with ours. Furthermore, in contrast to ours, most of these studies use a separate technique/tool for modeling the conceptual relations that need to be trained or applied separately with different mechanisms. We propose a lightweight, easy to use network structure for both coarse and fine classification with very high accuracy on large datasets.

It is important to distinguish our work from the  hyperbolic embedding studies~\cite{nickel17poincare,khrulkov20cvpr}. Khrulkov et al.~\cite{khrulkov20cvpr}, for example, attempt to compute an embedding where a more generic image reside closer to the origin than the specific ones. However, these papers do not describe -- and, it is neither obvious nor straightforward -- how to determine the concept classes from these embedded points as we propose in this paper. 


\section{Proposed Method}
Given an input image, the goal of our proposed method is to determine its category (leaf node in the hierarchy) \emph{and} a list of its concept superclasses (i.e., ancestors in the ontology). As an example, for an image of a Chimpanzee, the proposed algorithm produces predictions for 1) the category Chimpanzee, and 2) an ordered list of ancestor concepts: Living thing $\to$ Chordate $\to$ Mammal $\to$ Primate $\to$ Chimpanzee. We illustrate (and experiment with) the proposed model for image classification in this paper. 

Our CNN architecture is designed to encompass the chain of relationships among the category and the predecessor concepts in the dense layers. We utilize an existing label hierarchy/ontology to guide the design of the dense layers, but do not use the hierarchy in prediction.  In order to maximize the information within an ontology and to reduce the number of variables in the dense layers, we condense the original label hierarchy as explained in Section~\ref{S:CONCEPT_HIER}. In our design of multilayer dense connections, each concept is associated with a set of hidden nodes. These hidden nodes are in turn connected to those of its children (both category and concept) and the output prediction nodes. Section~\ref{S:ARCH} elaborates these connections and the loss functions to train the network.

\subsection{Condensed Concept Hierarchy}\label{S:CONCEPT_HIER}

In general, a concept class decomposes to multiple sub-concepts in an ontology, e.g., ImageNet12~\cite{imagenet} subset of WordNet~\cite{wordnet}. However, the lineage of a parent to a single child (e.g., Entity $\to$ Physical Entity $\to$ Abstraction ) is redundant and does not provide much information. Similarly, parent concepts with highly imbalanced distribution of descendants are not informative as well. Modeling the redundant and uninformative concepts will increase the network size with no information gain.

We reorganize the given ontology to reduce such redundancy. We assume the hierarchy to be a directed acyclic graph (DAG) and perform a depth first search (DFS) traversal on it. During the traversal, we first prune the label hierarchy based on the distribution of descendants of a concept node. Let $\eta_{\gamma}$ denote the number of all descendants of a concept indexed by $\gamma$. Any child node $\gamma_{\text{\tt{ch}}} \in \text{\tt{Children}}(\gamma)$  with  ${\eta_{\gamma_{\text{\tt{ch}}}} \over \eta_{\gamma}} \ge \tau$, i.e., the overlap between descendants $\gamma$ and $\gamma_{\text{\tt{ch}}}$  is  more than $\tau$, is absorbed by the parent $\gamma$. This process is applied recursively to yield a balanced distribution of descendants of any concept in the resulting hierarchy. In addition, we remove any concept $\gamma$ in the  structure with a descendant count $\eta_\gamma < \delta$ and append the children set $\text{\tt{Children}}(\gamma)$ to those of its parent $\gamma_{\text{\tt{pa}}}$. Conceptually, it is not worth modeling a concept node with only a few descendants.

We depict the differences between the original and modified ontologies in Figure~\ref{F:HIER}. Our strategy removes all the redundant single parent-child connections. Furthermore, the child concepts with imbalanced descendant counts (e.g., Physical Entity, Object etc.) are absorbed by their respective parents. As a result, the distributions of the descendants for children concepts are more balanced in the compressed version (right) than that in the original version (left).  As the network connections are dependent on the concept hierarchy, this reduction of redundancy in the ontologies is crucial for our method. It is worth noting that the proposed modification added direct concept-category relations in the middle layers of the hierarchy. 

Executing a DFS on a DAG ontology may lead to a few ambiguous grouping of few concepts and categories as \cite{deng14hex} pointed out. We adopted DFS for simplicity to compute the compressed graph, which can be replaced by an unambiguous one whenever it is available; the methods of~\cite{deng11labeltree,bengio10nips} can also be applied to generate the ontology.

\subsection{Network Architecture}\label{S:ARCH}
Our proposed algorithm aims to model the abridged label hierarchy with dense connections. As Figure~\ref{F:HIER} suggests, there are multiple kinds of dense connectivities in our proposed classification layer.  Each concept in the hierarchy corresponds to one set of hidden nodes that essentially represent the concept. These hidden nodes are connected to those representing its children, if any. For example, if Mammal, Bird and Reptile are the descendant concept of Chordate,  there will be all to all connections from the hidden nodes representing Chordate to those accounting for Mammal, Bird and Reptile. As such, the computation of the hidden representation of a child concept is conditioned upon that of its parent.

Given the representation captured in the hidden nodes, two types of output prediction nodes detect the presence of the concept itself and any children category in the input. An additional type of connectivity explicitly constrains the concept and category predictions to follow the hierarchical organization of the ontology. We illustrate each of these connections below.



\noindent \textbf{Modeling Concepts and Categories:} Let us denote by $z^\gamma$ and $\mathbf{h}^\gamma$ the output prediction variable  and the set of hidden nodes associated with the concept $\gamma$. The terms node and variables are used interchangeably in the description of our model.  Let concept $\gamma_{\text{\tt{ch}}}$ and category $j$ both be children of concept $\gamma$ in the hierarchy and $\mathbf{h}^{\gamma_{\text{\tt{ch}}}}$ and $x_j$ denote the hidden and the output prediction variables for them respectively. The proposed model computes the output prediction $z^\gamma$ and initial values  $\mathbf{\tilde{h}}^{\gamma_{\text{\tt{ch}}}}$, $\tilde{x}_j$ for quantities of the children concept and categories using the following dense connections.
\vspace{-0.2cm}
\begin{eqnarray}
\small
z^\gamma &=& \phi \Bigl (~\sum_i u_i^\gamma h_i^\gamma + b_z^\gamma~ \Bigr ) \nonumber \\ \tilde{x}_j &=& \psi \Bigl (~\sum_i v_{i,j}^\gamma ~h_i^\gamma + b_j^\gamma~ \Bigr ) \nonumber \\ \tilde{h}_{i'}^{\gamma_{\text{\tt{ch}}}} &=& \omega \Bigl (~\sum_i w_{i,i'}^\gamma ~h_i^\gamma + b_{i'}^\gamma~ \Bigr ) \label{E:DENSE_H}
\vspace{-0.9cm}
\end{eqnarray}
In these equations, $u, v, w$ and $b$ are the weights/biases of the dense connectivity and $h_i$ corresponds to the $i$-th value of $\mathbf{h}$. The activation functions $\phi, \psi, \omega$ utilized for these different quantities are $\phi = \text{Sigmoid}, ~\psi = \text{Identity}, ~\omega = \text{ReLU}$. In essence, $\mathbf{h}^\gamma$ encodes an internal description for concept $\gamma$ and $z^\gamma$ predicts the presence of it in the input image. The connections between $\mathbf{h}^\gamma$ and $\mathbf{h}^{\gamma_{\tt{ch}}}$ of its child concept enforces the descriptions for $\gamma_{\tt{ch}}$ to be derived from, and therefore dependent on, that of its parent.

In our design, the number $d^\gamma = |\mathbf{h}^\gamma|$ of nodes representing a concept $\gamma > 0$ is directly proportional to the total number of descendants $\eta_\gamma$ of $\gamma$. We have used $d^\gamma = \mu \eta_\gamma$ for this study with $\mu = 2$. The flattened output of the final feature layer of an existing network architecture (e.g., ResNet-50 or InceptionV4 etc.) is utilized to populate $\mathbf{h}^0$ and its size depends on the particular architecture used. We do not predict the root concept $z^0$ of the hierarchy (e.g., Entity for ImageNet12) since all categories descend from it.  

\begin{figure}
\vspace{-0.05in}
\begin{center}
\includegraphics[width=0.4\textwidth,height=0.41\textwidth]{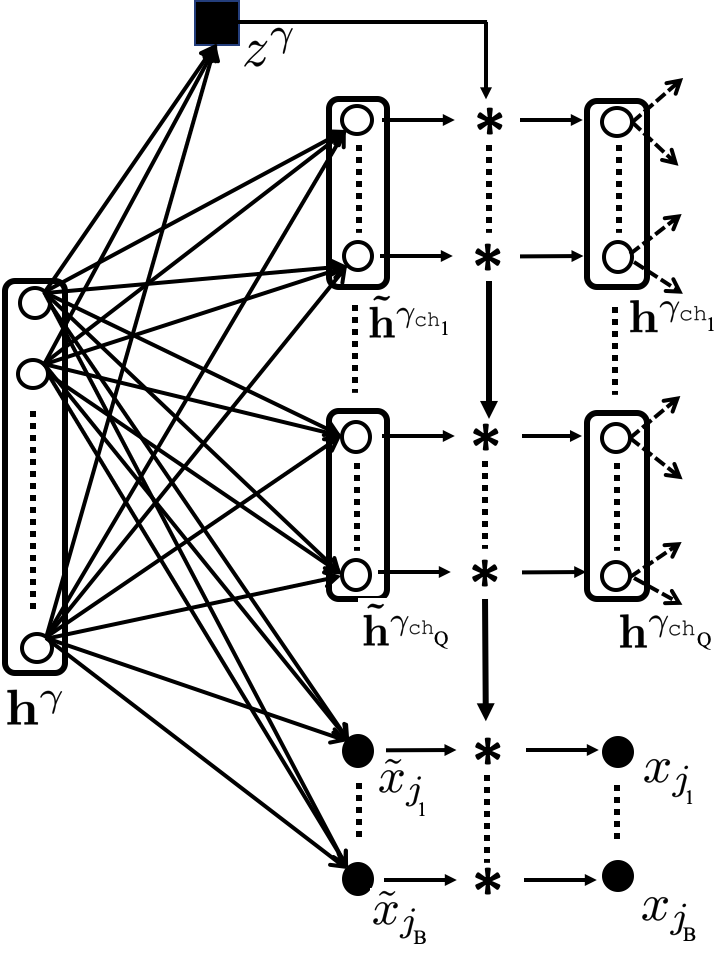}
\end{center}
\vspace{-0.7cm}
\caption{\small Schematic view of proposed dense connections. The solid square and circle nodes correspond to the concept and category prediction node respectively, whereas the empty circles depicts the hidden nodes. We assume the concept $\gamma$ has $Q$ concepts and $B$ categories as children.}
\label{F:DENSE}
\vspace{-0.5cm}
\end{figure}
\noindent \textbf{Concept Category Label Constraints:} The values for category prediction $x_j$ and hidden nodes for child concept $\mathbf{h}^{\gamma_{\text{\tt{ch}}}}$ are calculated by multiplying initial values of these quantities with the concept prediction $z^\gamma$.
\vspace{-0.2cm}
\begin{equation}
{x}_j  = \tilde{x}_j * z^\gamma; ~~
{h}_{i'}^{\gamma_{\text{\tt{ch}}}} = \tilde{h}_{i'}^{\gamma_{\text{\tt{ch}}}} * z^\gamma     
\vspace{-0.2cm}
\end{equation}

Note that, the Sigmoid activation constrains the value of $z^\gamma$ to be $z^\gamma \in [ 0, 1 ]$. In effect, the node $z^\gamma$ plays an excitatory or inhibitory role based on the predicted value of the concept $\gamma$. This constraint enforces that the nodes representing any child of concept $\gamma$, whether it is a category ($x_j$) or another downstream sub-concept ($\mathbf{h}^{\gamma_{\text{\tt{ch}}}}$), be activated only if the concept itself is correctly predicted.

The category predictions for an input image are computed by applying Softmax activation over all category nodes $\{x_1, x_2, \dots, x_{N}\}$, where $N$ is the total number of categories. The predictions for $M$ concepts are given by the collection of the variables $\{z^1, z^2, \dots, z^M\}$. The hierarchical relationship among the variables $\{z^1, z^2, \dots, z^M\}$ are enforced by construction. Observe that, while an image can be classified to only one (e.g., Chimpanzee) of the $N$ categories, multiple concepts (e.g., Primate, Mammal, Chordate) at different levels of the hierarchy can be set to 1. 


Figure~\ref{F:DENSE} clarifies the proposed dense arrangement between the hidden nodes $\mathbf{h}^\gamma$ and its children concept nodes  as well as the prediction outputs. The hidden nodes $\mathbf{h}^\gamma$ (shown in empty circles) are connected to those of its $Q$  children concepts and $B$ category output variables (solid circles) to compute the initial quantities $\{ \mathbf{\tilde{h}}^{\gamma_{\text{\tt{ch}}_1}}, \dots, \mathbf{\tilde{h}}^{\gamma_{\text{\tt{ch}}_Q}} \}$ and  $\{ \tilde{x}_{j_1}, \dots, \tilde{x}_{j_B} \}$ respectively. The concept prediction $z^\gamma$ (solid square) is computed by another dense connections which modulates the final values of $\{\mathbf{h}^{\gamma_{\text{\tt{ch}}_1}},  \dots,  \mathbf{h}^{\gamma_{\text{\tt{ch}}_Q}} \}$ and  $\{ x_{j_1}, \dots, x_{j_B} \}$ for the concept and category variables respectively via multiplication.

\noindent\textbf{Number of Variables: } The total number of weights in proposed multilayer dense connection with $\alpha$-way decomposition is $ \le \mu d^0 \bigl ( N + \rho + {\alpha \over (\alpha - 1)} \bigr )$ where $\rho$ is the height of the hierarchy and $\mu$ is the fixed multiplier used to set the number of hidden nodes for concept $\gamma$ (see supplementary material for details). In comparison, the final (single) layer of conventional CNN classifiers consists of $d^0 N$ variables.


\noindent \textbf{Loss Functions:} The proposed method minimizes (a weighted summation of) two different losses for the two types of output nodes. For the category predictions, we utilize a cross-entropy loss $L_{CE}(x)$ computed over the $N$ category labels and the network outputs $x_j,~ j=1, \dots, N$. 

The groundtruth label for $z^\gamma$ is a binary value quantifying the presence or absence of the concept $\gamma$ in the chain of superclasses. One could apply a Mean Squared Error (MSE) or a binary cross-entropy loss for each $z^\gamma$ to learn this multilabel classification. Experimentally we tested both and found that they are both effective for training the $z$ values, but the binary cross-entropy for each $z^\gamma$ leads to faster convergence that MSE. 


The proposed method minimizes the combined loss $L_{CE} + \lambda~L_{CON}$ with the balancing weight $\lambda$ fixed to $\lambda = 5$ in all our experiments. Note that, while the error for any category is backpropagated through its predecessor concepts due to the dependence imposed by construction, one needs to ensure that other concepts -- that are not related to the category -- to remain $0$ in the $\mathbf{z}$. This is exactly the constraint enforced by  $L_{CON}(z)$.


\section{Experiments \& Results}
The proposed architecture has been trained and tested on multiple datasets with different ontologies. For all experiments, we used two popular CNN architectures for feature computation layers,   ResNet-50~\cite{resnet,resnet2} and Inception V4~\cite{inceptionv3,inceptionv4}  (pretrained from~\cite{pretrained}) and replace the last dense layers with the proposed connections. 




\noindent \textbf{Inference:} During inference of the proposed network, we select the category with largest softmax probability for category classification as usual. The parent concepts are classified by thresholding $\mathbf{z}$ values. We also set any $z^\gamma = 0$ if the variable for its parent $z^{\gamma_{\text{\tt{pa}}}} = 0$ (i.e., lower than a confidence threshold). If more than one child of any concept is detected, we select the one with the highest confidence among them to compute the concept chain.  


\subsection{ImageNet12 Dataset Results}\label{S:IMGNET}

We utilize the ontology provided by the ImageNet12 dataset~\cite{imagenet} to design our dense layers. All the labels of ImageNet12 between $[1,1000]$ correspond to the $N=1000$ category classes and labels $>1000$ are assigned to the $860$ concept superclasses. After compressing the label hierarchy using the methods described in Section~\ref{S:CONCEPT_HIER} with $\tau = 90\%$, $\delta=20$, there are $40$ concept labels left in the hierarchy which has a height of $\rho = 7$. 

Adding the proposed multilayer dense connections increased the total number of variables of the ResNet-50 model by a factor of $1.182$ ($30.29 \over 25.61$M). For the InceptionV4 model, the increase is $1.059$ ($46.89 \over 44.24$M). This suggests that the increase induced by our proposed model is far lower than the analytical estimate (Section~\ref{S:ARCH}) in practice and is tolerable with respect to overall network size.

We demonstrate the performance of two variants of the proposed method that were trained on ImageNet12 training set and tested on its validation set. For one variant of our proposed CNN models, \textbf{MD-RN} with ResNet-50 and \textbf{MD-IC} with Inception V4 backbones, only the weights of the multilayer dense connections were trained; the weights of the feature layers were kept at their original pretrained values.  The other variant \textbf{MD-RN-FT} performs finetuning, i.e., it initiates the weights in the feature layer from the pretrained ResNet-50 and trains all variables in the network. 


\noindent\textbf{Baselines:} As the first baseline, we build a CNN classifier for $N+M$ classes with a single flat dense layer for each feature computation architecture. These baselines are trained with the same loss functions of the their multilayer counterparts. The flat dense connection baselines with ResNet-50, ResNet-50 Fintetuned and Inception are referred to as BL-RN, BL-RN-FT, BL-IC respectively. 

We utilize a modified version of the hedging method of Deng et al.~\cite{deng12hedge} as the second baseline. Our modified version uses ReasNet-50 and Inception V4 based classifiers instead of classical versions and are dubbed HG-RN and HG-IC respectively. We observed the optimal strategy to compute category and concept chain from the output of the hedge model (which is either category or max reward concept) is to use the category prediction as is and use the concept with maximum reward, along with all its ancestors, as the predicted concept chain. The hedge model utilizes the same underlying hierarchy as ours and we read off the prediction for concepts that overlap with our condensed hierarchy.

\noindent \textbf{Training:}
We trained MD-RN, MD-RN-FT and MD-IC following standard training procedures of ~\cite{inceptionv4,xception}. The details of training parameters and techniques are described in supplementary material for space limitation. The initial classifiers for HG-RN and HG-IC were trained from scratch on 93\% of training examples for each class in ImageNet 2012 training set (achieving 74\% and 76\% category accuracy on validation set respectively). The remaining 7\% examples were used to estimate the dual variable maximizing the reward for multiple accuracy guarantees. The results reported here correspond to the dual variable, accuracy guarantee with best overall $Acc_{\text{COMB}}$.
\begin{table}
\vspace{-0.2in}
\caption{\small Accuracy comparison on ImageNet12 val set (single crop top-1). The proposed models achieve significantly higher $Acc_{\text{CON}}, Acc_{\text{COMB}}$ than the baselines.}
\vspace{-0.4cm}
\begin{center}
\scriptsize
\begin{tabular}{|c|c|c|c|c|c|}
\hline
Method & $Acc_{\text{CAT}}$ & $Acc_{\text{CON}}$ &$Acc_{\text{COMB}}$& mhP & mhR\\
\hline
HG(Deng etal.12)-RN& 68.97 & 82.9 & 66.15 & 93.28 & 94.57 \\
BL-RN& 76.26 & 62.66 &53.01&92.36 & 89.48\\
BL-RN-FT& 76.0 & 69.04 & 58.89 & 93.91 & 91.06 \\
\hline
MD-RN(ours)& 75.94 & 80.25 &69.05& 94.22 & 93.68 \\
MD-RN-FT(ours)& 75.91 & 82.36 &\textbf{70.45}& 93.3 & 94.75\\ 
\hline
\hline
HG(Deng etal.12)-IC& 71.49 & 84.28 &68.87& 93.75& 95.07\\
BL-IC& 80.11 & 72.68 &63.56 & 95.03 & 91.51\\ 
\hline
MD-IC(ours)& 80.02 & 88.18 &\textbf{77.76} &95.46 & 96.1\\ 
\hline
\end{tabular}
\end{center}
\vspace{-0.3in}
\label{T:RESULT_FT}
\end{table}

\noindent \textbf{Evaluation:} We report the single crop top-1 accuracy results $Acc_{\text{CAT}}$ for categories when they are available. For concept inference, we report two measures based on hierarchical precision  and recall~\cite{costa2007ARO}, which computes the precision an recall between the predicted and true chain of concepts :  1) the mean hierarchical precision (mhP) and recall (mhR) over all images of test set, and, 2) percentage of images $Acc_{\text{CON}}$ that were predicted with 100\% hP and hR. The combined accuracy, $Acc_{\text{COMB}}$, denotes the percentage of images with both $Acc_{\text{CAT}} = Acc_{\text{CON}} = 1$.

Table~\ref{T:RESULT_FT} reports the accuracies for all the aforementioned models. Table~\ref{T:RESULT_FT} demonstrates that the MD-RN, MD-RN-FT and MD-IC networks with proposed multilayer dense connections achieved a comparable category accuracy $ACC_{\text{CAT}}$ of the flat baselines BL-RN, BL-RN-FT and BL-IC (which are similar to the published results).  The category accuracy of HG-RN, HG-IC are significantly lower than the aforementioned models -- a phenomenon also observed by~\cite{deng12hedge}. This is due to the inherent design of the hedge model to only predict concept classes (and not the leaf categories) for some examples.

Table~\ref{T:RESULT_FT} also suggests that an unconstrained flat dense connection is insufficient for learning the superclass sequence. The flat baseline methods BL-RN, BL-RN-FT and BL-IC attempt to compute the concept classes independently using the same single layer dense connections that are utilized for category classification. While these single layer connections can classify the categories very accurately, the concept accuracies $Acc_{\text{CON}}$ of BL-RN, BL-RN-FT and BL-IC are approximately $7 \sim 14\%$ lower than category accuracies $Acc_{\text{CAT}}$. This suggests that the feature representations captured (in the last convolutional layer) by ResNet-50 and Inception V4 alone are not as informative for concept prediction as they are for category classification. Our multilayer hierarchical design combines these initial representations to yield a refined encoding in the hidden variables $h^\gamma$ to improve the concept prediction accuracy $Acc_{\text{CON}}$ by $>13\%$. 

The comparison also implies that the proposed top-down hierarchical constraints on category classification to preserve the ancestor concept chain is superior to the hedging technique~\cite{deng12hedge} of concept prediction distilled from the aggregated probabilities accumulated bottom-up from category confidence. As we will show in next section, a top-down constrained prediction is more robust to the challenging examples because it does not rely primarily on the category classification probabilities. 

Overall, the baseline methods can  learn to classify either the finer categories or the concept chain well but the proposed CNN architecture is capable of classifying both simultaneously with high accuracy. The proposed models MD-RN, MD-RN-FT and MD-IC achieves a significantly higher overall accuracy $Acc_{\text{COMB}}$ than both the baselines. The inception based model MD-IC is also more accurate than HG-IC for concept prediction -- indeed, MD-IC leads to the best overall accuracy $Acc_{\text{COMB}}$ among all models.

\noindent \textbf{Analysis:} We provide ablation studies and analysis of the distribution of category classifications in supplementary material. 

\subsection{Performance on Naturally Adversarial Images}\label{S:IMGNET-ADV}
Our next experiment demonstrates the robustness of the proposed network on challenging naturally adversarial image dataset ImageNet-A~\cite{hendrycks19adv}. Popular CNN architectures (e.g. ResNet) trained on standard datasets classify images from ImageNet-A to widely different semantic categories. It is logical to speculate that algorithms (e.g., \cite{deng12hedge}) that compute marginal probabilities through bottom up aggregation will be less effective for determining the ancestor concepts given these widely divergent predictions.  There are 7500 images in ImageNet-A from 200 categories. The ResNet-50 model we downloaded from the tensorflow website (pretrained on full ImageNet 2012) could classify only 0.73\% of the images. Since these 200 categories are a subset of the 1000 categories of ImageNet 2012 dataset, one can use the same concept structure for hierarchical classification.

\begin{table}[t]
\vspace{-0.2in}
\caption{\small Accuracy comparison on ImageNet-A (single crop top 1). The proposed models achieve significantly higher $Acc_{\text{CON}}, Acc_{\text{COMB}}$ than the baselines.}
\vspace{-0.4cm}
\begin{center}
\scriptsize
\begin{tabular}{|c|c|c|c|c|c|}
\hline
Method & $Acc_{\text{CAT}}$ & $Acc_{\text{CON}}$ &$Acc_{\text{COMB}}$& mhP & mhR\\
\hline
HG(Deng etal.12)-RN& 0.37 & 14.36 & 0.3 & 57.08 & 55.99 \\
BL-RN-FT& 0.86 & 11.52 & 0.26 & 59.12 & 54.33 \\
MD-RN-FT (ours)& 1.06 & \textbf{16.13} & 0.77 & 57.6 & 58.0 \\ 
\hline
HG(Deng etal.12)-IC& 3.18 & 19.76 & 2.8 & 59.73 & 58.77 \\
BL-IC& 7.54 & 16.62 & 3.56 & 61.78 & 58.07\\
MD-IC (ours)& 7.1 & \textbf{23.5} & 6.1 & 63.79 & 61.46 \\ 
\hline
\end{tabular}
\end{center}
\vspace{-0.3in}
\label{T:RESULT_ADV}
\end{table}

\begin{figure*}[t]
\vspace{-0.2in}
\begin{center}
\includegraphics[width=0.97\textwidth,height=0.225\textwidth]{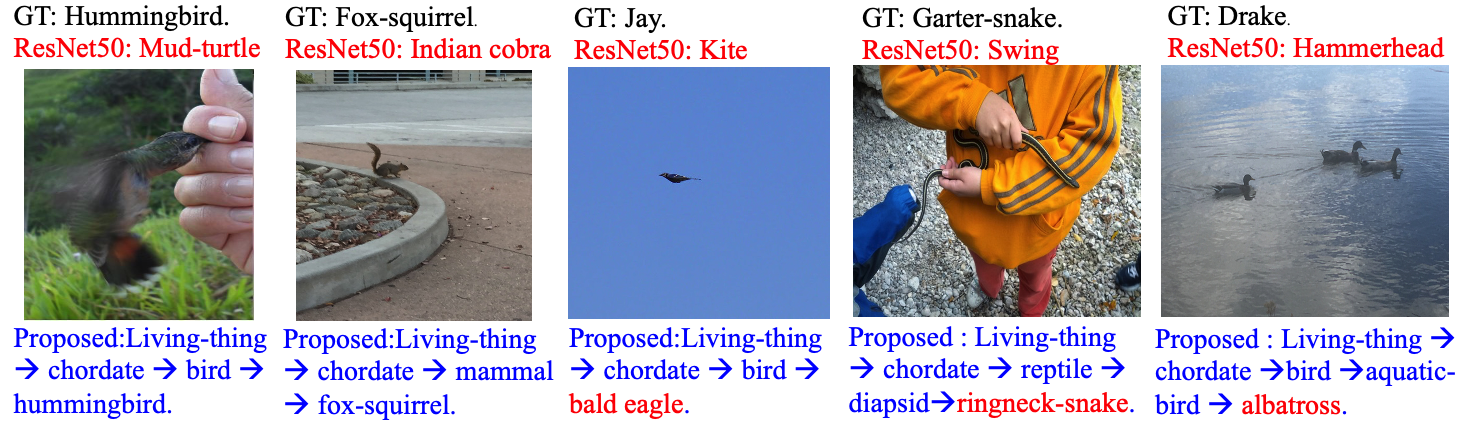}
\end{center}
\vspace{-0.4cm}
\caption{\small Sample comparison between proposed MD-RN \& pretrained ResNet50 outputs on naturally adversarial images~\cite{hendrycks19adv}.}
\label{F:RESULT_ADV_QUAL}
\vspace{-0.1cm}
\end{figure*}

\begin{table*}[t]
\vspace{-0.1cm}
\caption{\small Concept accuracy comparison PASCAL VOC 2012 trainval subset.  The proposed method achieves significantly higher $Acc_{\text{CON}}$ than the baseline for both backbones.}
\scriptsize
\vspace{-0.3cm}
\begin{center}
\begin{tabular}{|c|c|c|c|c|c|c|c|c|c|c|c|c|c|c|c|c|c|}
\hline
Method&cat&cow&horse&sheep&plane&boat&bike&mbike&bus&train&car&botl&chair&dtable&sofa&montr&avg\\
\hline
PrAggRN&71.2&73.1&51.4&72.1&75.1&55.3&73.4&34.1&61.7&4.4&2.4&36&29.3&8.5&30.8&37.8&46.2\\
BL-RN-FT&57.2&53.9&43&53.8&72.5&50.2&65.7&35.3&56.8&1.5&1.6&21.6&17.9&1&13.1&30.7&38.06\\
\hline
MD-RN-FT&70.7&78&56.2&70.8&85.8&69.1&76.8&4.9&54.6&3.6&2.7&43.4&36.2&13.8&44.7&53&\textbf{48.84}\\ 

\hline
\hline
PrAggIC&69.8&80.7&59.8&80.7&80.3&59.7&75.4&30.1&64.6&4&2&38.9&31.8&9.5&36.4&33.3&48.32\\
BL-IC&64.9&63.2&47.2&60.5&80.3&45.9&45&2.9&71&1.9&2.1&25.3&19.3&3.1&24&32.3&40.3\\
\hline
MD-IC&74.6&84.1&60.9&80.7&86&63.1&77.5&8.1&65.8&3.8&1.7&46.4&39.4&15.9&40.9&43.3&\textbf{50.38}\\ 
\hline
\end{tabular}
\end{center}
\vspace{-0.6cm}
\label{T:RESULT_VOC}
\end{table*}

Table~\ref{T:RESULT_ADV} reports the quantitative performances of the  variants of proposed architecture as well as the baselines trained on ImageNet12 but applied to ImageNet-A. In addition,  Figure~\ref{F:RESULT_ADV_QUAL} compares sample predictions from the proposed MD-RN-FT network to those from the pretrained ResNet-50. One can observe that, ResNet-50 can  classify the images to semantically very different categories while the constraints in the proposed architecture enforces it to determine the ancestor superclasses correctly even when the leaf category is misclassified. These qualitative outputs help us elucidate the better performances of proposed  MD-RN-FT and MD-IC for concept prediction than the hedging technique~\cite{deng12hedge} that uses bottom up probability aggregation.



\subsection{Zero shot Concept Prediction on PASCAL VOC12}\label{S:VOC}
In this section, we test the capability of the proposed model to generalize the knowledge learned in the multilayer dense architecture. Ideally, the proposed network should be able to extrapolate its understanding of the concept superclasses learned from one set of categories to previously unseen categories. That is, after learning that a zebra is a mammal, it should be able to identify a horse (of any color) as a mammal too. 

To test generalization capability, we test classifiers learned from the ImageNet12 dataset on the trainval split of PASCAL VOC dataset~\cite{Everingham10} to detect coarse classes. Note that, there is no one to one correspondence between the categories of ImageNet and VOC~\cite{imgnetvoc}. We have excluded the images of \{Person, Pottedplant\} and \{Dog, Bird\} subsets since these categories were underrepresented overrepresented respectively in the ImageNet12 dataset. Removing these images leave us  6941 images from 16 categories. Each of the VOC12 categories were assigned to one of the coarse superclasses in the condensed label hierarchy of ImageNet dataset.

Because we are not utilizing VOC12 categories, we modify our second baseline to compute only the concept classes by thresholding the marginal probabilities as computed in~\cite{deng12hedge}. These are referred to as  PrAggRN and PrAggIC when computed from the initial ResNet-50 and Inception V4 classifiers respectively. The accuracy values for concept classification on VOC 2012 summarized in Table~\ref{T:RESULT_VOC}  clearly indicate the superiority of the proposed architecture to generalize the knowledge it learned from ImageNet12 hierarchy of ancestor superclasses. All the CNN models  resulted in weak performances of categories \{Train, Car, Diningtable\}. Some of these categories are associated with equivocal ancestry in the condensed ontology. For the category Train, the proposed MD-IC predicted Entity$\to$ Artifact$\to$ Instrumentality$\to$ Container$\to$ Wheeled vehicle $\to$ Self propelled vehicle for 56\% of images whereas the compressed ontology assigns it  the concept order : Entity$\to$ Artifact$\to$ Instrumentality$\to$ Conveyance. It is important to note that the performances of both the proposed method and baseline suffer due to the ambiguity -- i.e., the proposed method does not gain any unfair advantage on these categories.


\subsection{AwA2 Dataset Results}\label{S:AWA}
Our next experiment compares the performances of structural inference model of Hu et al.~\cite{hu16cvpr} and the HDCNN model of \cite{yan15hdcnn} with the proposed architecture on the Animals with Attributes (AwA2) dataset~\cite{AwA}. AwA2 dataset provides the attribute and category labels for 37322 images from 50 animal classes. We also adopt the same 28 superclasses and ontology used in \cite{hu16cvpr}  as is (no compression) for concept hierarchy. Note that, this hierarchy is slightly different from the ImageNet12 ontology.

We use a 60-40\% split of the images for train/test data as~\cite{hu16cvpr} suggests and report the average$\pm$std dev of the multiclass accuracy $Acc_{\text{CAT}}$ for category and the IoU acc for concept and attribute labels from 3 trials. Table~\ref{T:RESULT_AWA} compares the performances of the proposed MD-RN with that of \cite{hu16cvpr} (values taken from the paper) and the flat baseline BL-RN with ResNet-50 backbone. To determine the attributes, we extend our proposed architecture by adding a dense connection between concept and category prediction variables to the 85 attribute classes. The flat baseline BL-RN predicts the attributes classes independently in addition to animal and concept labels.

Following ~\cite{yan15hdcnn}, our implementation of the HDCNN model uses the last residual block (conv5\_x) of ResNet-50 along with the subsequent flat dense connections in independent blocks for leaf category and ancestor concepts whereas the remainder of the network was exploited as shared layers. The size of the resulting HDCNN network grows to 773M, roughly 25 fold larger than MD-RN. The probabilities of the leaf categories were computed by the weighted average of those of the immediate parents -- the same way ~\cite{yan15hdcnn} computes it. We have only trained the variables in the independent blocks (for a fair comparison with  MD-RN) by minimizing the same loss function as described in Section~\ref{S:ARCH}.
\begin{table}[t]
\vspace{-0.2in}
\caption{\small Accuracy comparison on AwA2 dataset for simultaneous category, concept, attribute classification.}
\scriptsize
\vspace{-0.25cm}
\begin{center}
\begin{tabular}{|c|c|c|c|}
\hline
Method& $Acc_{\text{CAT}}$& IoU Concept & IoU Attribute \\
\hline
Hu et al.~\cite{hu16cvpr}& $79.36\pm0.43$&$84.47\pm0.38$ & $86.92\pm0.18$\\
Yan et al.~\cite{yan15hdcnn}& $89.15\pm0.1$&$41.37\pm0.27$ &  -- \\
BL-RN&$92.48\pm0.05$&$93.66\pm0.42$&$89.08\pm0.01$\\
MD-RN (ours)&$92.16\pm0.17$&$\mathbf{97.01\pm0.05}$& $\mathbf{94.1\pm0.37}$\\
\hline
\end{tabular}
\end{center}
\vspace{-0.3in}
\label{T:RESULT_AWA}
\end{table}

Hu et al.~\cite{hu16cvpr} use a weaker CNN base (\cite{alexnet}) than ours, which is perhaps a factor leading to its inferior performance. The comparison suggests that, given CNN backbone with enhanced capacity, one can attain highly correct category, concept and attribute classification from our multilayer dense connection without relying on an additional complex tool, e.g., RNN. However, the lower IoU concept, IoU attribute of the flat baseline method attest that the improved performance of MD-RN does not arise solely from the ResNet-50 features and  strengthen our claim that the proposed mulitlayer connectivity amplifies the capacity of the feature representation. From the accuracy values, we also speculate that HDCNN approach perhaps can learn the leaf category and immediate parent concepts given its shallow hierarchy. But, to handle a multilevel hierarchy it needs far more complex structure and probably even more complex techniques to train it.

\subsection{Better mistakes on iNaturalist Dataset}\label{S:INAT}
We evaluated our model on the iNaturalist dataset~\cite{inaturalist} in order to compare with the hierarchical cross entropy technique of~\cite{bertinetto20cvpr}. This dataset contains 268,243 images from 1010 categories and 180 ancestor concepts. After compressing the ontology with $\tau = 100\%$, $\delta=20$, our hierarchy reduced to 6 levels with 40 superclasses. Following~\cite{bertinetto20cvpr}, we use a 70-30\% split for train-test and report results from 3 trials.  For proposed model, we trained the finetuned ResNet-50 variant MD-RN-FT adhering to the training strategy and parameters described in Section~\ref{S:IMGNET}. The baseline  ResNet-50 of was trained with hierarchical entropy (HXE) with $\alpha=0.1$ that yields the best performance~\cite{bertinetto20cvpr}. We also enlarge the input image (for both networks) to $256\times256$ to improve accuracy.

For a direct comparison, we report the performances in the evaluation measures used in~\cite{bertinetto20cvpr}. One of the evaluation criteria was $Acc_\text{CAT}$. The other measure is the mean height $h_\text{LCA}$ of the least common ancestor(LCA) between the \emph{groundtruth} ancestor concept chain of the category prediction and the true label. However, unlike~\cite{bertinetto20cvpr}, we compute $h_\text{LCA}$ for \emph{only the misclassified samples}  which, we believe to be more meaningful than $h_\text{LCA}$  for all examples. We also report the percentage $N_\text{diff}$ of examples for which the groundtruth concept chain differs from that of the predicted category. Note that $N_\text{diff}$ can be different from $Acc_\text{CAT}$ because a misclassification to a different finer category under same parent will not change $N_\text{diff}$.
\begin{table}[t]
\vspace{-0.2cm}
\caption{\small Single crop top 1 accuracy on iNaturalist images. Proposed method makes less severe  mistakes than ~\cite{bertinetto20cvpr}. Higher $Acc_\text{CAT}$  and lower $N_\text{diff}, h_\text{LCA}$ are better.}
\scriptsize
\vspace{-0.15cm}
\begin{center}
\begin{tabular}{|c|c|c|c|}
\hline
Method& $Acc_{\text{CAT}}$& $N_\text{diff}$ & $h_{\text{LCA}}$ \\
\hline
Bertinetto et al.\cite{bertinetto20cvpr}& $70.9\pm0.5$&$10.1\pm0.2$ & $3.33\pm0.005$\\
MD-RN-FT (ours)&$\mathbf{71.5\pm0.7}$&$\mathbf{8.6\pm0.4}$& $\mathbf{3.1\pm0.001}$\\
\hline
\end{tabular}
\end{center}
\vspace{-0.3in}
\label{T:RESULT_INAT}
\end{table}

As Table~\ref{T:RESULT_INAT} shows, a lower $N_\text{diff}$ of the proposed model implies it generates fewer mistakes in terms of true concept chain than~\cite{bertinetto20cvpr}. In addition, these mistakes are also less severe since the height $h_\text{LCA}$ of LCA is also smaller on average than that of ~\cite{bertinetto20cvpr}. The results suggest that constraining the variables $x, z$ to follow the hierarchy leads to more meaningful classification than hierarchical cross entropy training.   

\section{Discussion}\label{S:DISCUSSION}

In this paper, we present an effective CNN architecture with multilayer dense connections for integrated classification of coarse concepts and fine categoery classes. Our mulitlayer connections can conveniently be incorporated in a CNN model by replacing the single flat layer at the end and can be learned with standard training approaches --  often without training the whole network. We demonstrate the practical efficacy and robustness of the proposed architecture with different backbones on multiple datasets.


Our experiments demonstrate the advantage of the proposed architecture over 1) multinomial regression (flat connections) that exploits the learned feature representation of pretrained CNNs;  2) method utilizing marginal probabilities accumulated in a bottom up fashion from CNN output given an offline hierarchy~\cite{deng12hedge} and 3) existing techniques for superclass prediction that employ larger more complex models~\cite{yan15hdcnn, hu16cvpr}. Our top-down constraints enforces the category classification to conform to the semantically meaningful lineage of ancestor classes. This property not only allows zero-shot concept classification on unseen categories but also leads significantly better mistakes than~\cite{bertinetto20cvpr}. We believe this study will encourage researchers to adopt and improve methods for simultaneous detection of category and concepts.





{\small
\bibliographystyle{ieee_fullname}
\bibliography{iccv2021_concept}

\begin{thebibliography}{10}\itemsep=-1pt

\bibitem{pretrained}
\url{https://github.com/tensorflow/models/tree/master/research/slim.}

\bibitem{imgnetvoc}
\url{http://image-net.org/challenges/LSVRC/2012/analysis/}.

\bibitem{bengio10nips}
Samy Bengio, Jason Weston, and David Grangier.
\newblock Label embedding trees for large multi-class tasks.
\newblock In {\em NIPS}. 2010.

\bibitem{bertinetto20cvpr}
Luca Bertinetto, Romain Mueller, Konstantinos Tertikas, Sina Samangooei, and
  Nicholas~A. Lord.
\newblock Making better mistakes: Leveraging class hierarchies with deep
  networks.
\newblock In {\em Proceedings of the IEEE/CVF Conference on Computer Vision and
  Pattern Recognition (CVPR)}, June 2020.

\bibitem{cao18vqr}
Qingxing Cao, Xiaodan Liang, Bailing Li, Guanbin Li, and Liang Lin.
\newblock Visual question reasoning on general dependency tree.
\newblock {\em CoRR}, abs/1804.00105, 2018.

\bibitem{liang18deeplabv3}
Liang{-}Chieh Chen, Yukun Zhu, George Papandreou, Florian Schroff, and Hartwig
  Adam.
\newblock Encoder-decoder with atrous separable convolution for semantic image
  segmentation.
\newblock {\em CoRR}, abs/1802.02611, 2018.

\bibitem{xception}
François Chollet.
\newblock Xception: Deep learning with depthwise separable convolutions.
\newblock In {\em 2017 IEEE Conference on Computer Vision and Pattern
  Recognition (CVPR)}, pages 1800--1807, 2016.

\bibitem{costa2007ARO}
Eduardo~P. Costa, A. Lorena, A. Carvalho, and A. Freitas.
\newblock A review of performance evaluation measures for hierarchical
  classifiers.
\newblock In {\em {AAAI}}, 2007.

\bibitem{deformable}
J. {Dai}, H. {Qi}, Y. {Xiong}, Y. {Li}, G. {Zhang}, H. {Hu}, and Y. {Wei}.
\newblock Deformable convolutional networks.
\newblock In {\em 2017 IEEE International Conference on Computer Vision
  (ICCV)}, 2017.

\bibitem{deng10K}
Jia Deng, Alexander~C. Berg, Kai Li, and Li Fei-Fei.
\newblock What does classifying more than 10,000 image categories tell us?
\newblock In {\em {ECCV}}, 2010.

\bibitem{deng14hex}
Jia Deng, Nan Ding, Yangqing Jia, Andrea Frome, Kevin Murphy, Samy Bengio, Yuan
  Li, Hartmut Neven, and Hartwig Adam.
\newblock Large-scale object classification using label relation graphs.
\newblock In {\em ECCV}, 2014.

\bibitem{imagenet}
J. {Deng}, W. {Dong}, R. {Socher}, L. {Li}, {Kai Li}, and {Li Fei-Fei}.
\newblock Imagenet: A large-scale hierarchical image database.
\newblock In {\em {CVPR}}, 2009.

\bibitem{deng12hedge}
Jia Deng, Jonathan Krause, Alex Berg, and Li Fei-Fei.
\newblock Hedging your bets: Optimizing accuracy-specificity trade-offs in
  large scale visual recognition.
\newblock In {\em {CVPR}}, 2012.

\bibitem{deng11labeltree}
Jia Deng, Sanjeev Satheesh, Alexander~C. Berg, and Fei Li.
\newblock Fast and balanced: Efficient label tree learning for large scale
  object recognition.
\newblock In {\em NIPS}, 2011.

\bibitem{ding15phex}
Nan Ding, Jia Deng, Kevin Murphy, and Hartmut Neven.
\newblock Probabilistic label relation graphs with ising models.
\newblock In {\em {ICCV}}, 2015.

\bibitem{Everingham10}
M. Everingham, L. Van~Gool, C.~K.~I. Williams, J. Winn, and A. Zisserman.
\newblock The {Pascal} visual object classes ({VOC}) challenge.
\newblock {\em International Journal of Computer Vision}, 88(2):303--338, June
  2010.

\bibitem{wordnet}
Christiane Fellbaum, editor.
\newblock {\em WordNet: An Electronic Lexical Database}.
\newblock Language, Speech, and Communication. MIT Press, Cambridge, MA, 1998.

\bibitem{fergus2010SemanticLS}
Rob Fergus, Hector Bernal, Yair Weiss, and Antonio Torralba.
\newblock Semantic label sharing for learning with many categories.
\newblock In {\em ECCV}, 2010.

\bibitem{Guo2017CNNRNNAL}
Yanming Guo, Yu Liu, Erwin~M. Bakker, Yuanhao Guo, and Michael~S. Lew.
\newblock Cnn-rnn: a large-scale hierarchical image classification framework.
\newblock {\em Multimedia Tools and Applications}, 77:10251--10271, 2017.

\bibitem{mask-rcnn}
K. {He}, G. {Gkioxari}, P. {Dollár}, and R. {Girshick}.
\newblock Mask r-cnn.
\newblock In {\em {ICCV}}, pages 2980--2988, 2017.

\bibitem{resnet}
Kaiming He, Xiangyu Zhang, Shaoqing Ren, and Jian Sun.
\newblock Deep residual learning for image recognition.
\newblock {\em CoRR}, abs/1512.03385, 2015.

\bibitem{resnet2}
Kaiming He, Xiangyu Zhang, Shaoqing Ren, and Jian Sun.
\newblock Identity mappings in deep residual networks.
\newblock {\em CoRR}, abs/1603.05027, 2016.

\bibitem{hendrycks19adv}
Dan Hendrycks, Kevin Zhao, Steven Basart, Jacob Steinhardt, and Dawn Song.
\newblock Natural adversarial examples.
\newblock {\em CoRR}, abs/1907.07174, 2019.

\bibitem{hu16cvpr}
H. {Hu}, G. {Zhou}, Z. {Deng}, Z. {Liao}, and G. {Mori}.
\newblock Learning structured inference neural networks with label relations.
\newblock In {\em 2016 IEEE Conference on Computer Vision and Pattern
  Recognition (CVPR)}, 2016.

\bibitem{squeezenet}
J. {Hu}, L. {Shen}, and G. {Sun}.
\newblock Squeeze-and-excitation networks.
\newblock In {\em 2018 IEEE/CVF Conference on Computer Vision and Pattern
  Recognition}, 2018.

\bibitem{densenet}
G. {Huang}, Z. {Liu}, L. v.~d. {Maaten}, and K.~Q. {Weinberger}.
\newblock Densely connected convolutional networks.
\newblock In {\em 2017 IEEE Conference on Computer Vision and Pattern
  Recognition (CVPR)}, 2017.

\bibitem{khrulkov20cvpr}
Valentin Khrulkov, Leyla Mirvakhabova, Evgeniya Ustinova, Ivan Oseledets, and
  Victor Lempitsky.
\newblock Hyperbolic image embeddings.
\newblock In {\em {CVPR}}, 2020.

\bibitem{alexnet}
Alex Krizhevsky, Ilya Sutskever, and Geoffrey~E Hinton.
\newblock Imagenet classification with deep convolutional neural networks.
\newblock In {\em Advances in Neural Information Processing Systems 25}. 2012.

\bibitem{AwA}
Christoph~H. Lampert, Hannes Nickisch, and Stefan Harmeling.
\newblock Attribute-based classification for zero-shot visual object
  categorization.
\newblock 36(3):453–465, 2014.

\bibitem{lin2017focal}
Tsung-Yi Lin, Priya Goyal, Ross~B. Girshick, Kaiming He, and Piotr Doll{\'a}r.
\newblock Focal loss for dense object detection.
\newblock In {\em {ICCV}}, 2017.

\bibitem{ssd}
Wei Liu, Dragomir Anguelov, Dumitru Erhan, Christian Szegedy, Scott~E. Reed,
  Cheng{-}Yang Fu, and Alexander~C. Berg.
\newblock {SSD:} single shot multibox detector.
\newblock {\em CoRR}, abs/1512.02325, 2015.

\bibitem{nickel17poincare}
Maximillian Nickel and Douwe Kiela.
\newblock Poincar\'{e} embeddings for learning hierarchical representations.
\newblock In {\em Advances in Neural Information Processing Systems}, 2017.

\bibitem{vgg}
K. Simonyan and A. Zisserman.
\newblock Very deep convolutional networks for large-scale image recognition.
\newblock In {\em {ICLR}}, 2015.

\bibitem{inceptionv4}
Christian Szegedy, Sergey Ioffe, Vincent Vanhoucke, and Alex~A. Alemi.
\newblock Inception-v4, inception-resnet and the impact of residual connections
  on learning.
\newblock In {\em ICLR 2016 Workshop}, 2016.

\bibitem{inceptionv3}
Christian Szegedy, Vincent Vanhoucke, Sergey Ioffe, Jonathon Shlens, and
  Zbigniew Wojna.
\newblock Rethinking the inception architecture for computer vision.
\newblock {\em CoRR}, abs/1512.00567, 2015.

\bibitem{efficientnet}
Mingxing Tan and Quoc Le.
\newblock {E}fficient{N}et: Rethinking model scaling for convolutional neural
  networks.
\newblock In {\em ICML}, 2019.

\bibitem{torralba04boosting}
A. {Torralba}, K.~P. {Murphy}, and W.~T. {Freeman}.
\newblock Sharing features: efficient boosting procedures for multiclass object
  detection.
\newblock In {\em {CVPR}}, 2004.

\bibitem{inaturalist}
G. {Van Horn}, O. {Mac Aodha}, Y. {Song}, Y. {Cui}, C. {Sun}, A. {Shepard}, H.
  {Adam}, P. {Perona}, and S. {Belongie}.
\newblock The inaturalist species classification and detection dataset.
\newblock In {\em 2018 IEEE/CVF Conference on Computer Vision and Pattern
  Recognition}, 2018.

\bibitem{fvqa}
Peng Wang, Qi Wu, Chunhua Shen, Anton van~den Hengel, and Anthony~R. Dick.
\newblock {FVQA:} fact-based visual question answering.
\newblock {\em CoRR}, abs/1606.05433, 2016.

\bibitem{wu04cascade}
Jianxin Wu, James~M Rehg, and Matthew~D. Mullin.
\newblock Learning a rare event detection cascade by direct feature selection.
\newblock In {\em {NIPS}}. 2004.

\bibitem{resnetxt}
S. {Xie}, R. {Girshick}, P. {Dollár}, Z. {Tu}, and K. {He}.
\newblock Aggregated residual transformations for deep neural networks.
\newblock In {\em 2017 IEEE Conference on Computer Vision and Pattern
  Recognition (CVPR)}, 2017.

\bibitem{yan15hdcnn}
Zhicheng Yan, Hao Zhang, Robinson Piramuthu, Vignesh Jagadeesh, Dennis DeCoste,
  Wei Di, and Yizhou Yu.
\newblock {HD-CNN}: Hierarchical deep convolutional neural networks for large
  scale visual recognition.
\newblock In {\em ICCV}, pages 2740--2748, 2015.

\bibitem{zhao11largescale}
Bin Zhao, Fei Li, and Eric~P. Xing.
\newblock Large-scale category structure aware image categorization.
\newblock In {\em NIPS}. 2011.

\end{thebibliography}
}

\section{Supplementary Material}

\subsection{Number of Variables in Multilayer Dense Connections}\label{A:NUM_VAR}
This section quantifies the increase in the number of weights in the dense layers induced by our multilayer computations. The CNN classifiers typically consists of $d^0 N$ connections in dense layer, where $d^0$ and $N$ are the size of the last feature layer and  number of category classes respectively. In proposed multilayer dense connections with a balanced $\alpha$-way decomposition of the concepts, the total number of weights is $ \le \mu d^0 \Bigl ( N + \rho + {\alpha \over (\alpha - 1)} \Bigr )$ where $\rho$ is the height of the hierarchy and $\mu$ is the fixed multiplier used to set the number of hidden nodes for concept $\gamma$.

In a balanced decomposition, the number of hidden nodes for concepts reduces by a constant factor $\alpha$. With the size of smallest set of hidden nodes as $\delta$, the max layer of the dense connections is $ \rho = \log_\alpha ({d^0 \over \delta})$. For any concept prediction $z^\gamma$, we need $d^\gamma = {\mu d^0 \over \alpha^l }$ weights at layer $l$ i.e., max number of weights for concept class prediction is $\mu d^0 \rho = \mu d^0 \log_\alpha ({d^0 \over \delta})$. The number of concept-concept connections can be calculated as $d^0({\mu \over \alpha} + {\mu \over \alpha^2}+ \dots + {\mu \over \alpha^\rho}) = \mu d^0 {(1 - {1 \over \alpha^\rho}) \over (1 - {1 \over \alpha})}  \le \mu d^0 { 1 \over (1 - {1 \over \alpha})}$. 

In order to predict a category variable $x_j \in \text{\tt{Children}}(\gamma)$, $d^\gamma$ weights are necessary. Since any $\mu d^0 \ge d^\gamma = \mu {d^0 \over \alpha^l}$ at any level $l$ in a balanced decomposition, the total number of weights for category class prediction must be $\le \mu d^0 N $.

\subsection{CNN Training Details} \label{A:TRN_DETAILS}
For MD-RN, MD-RN-FT and MD-IC training, we used the RMSProp optimizer with momentum in our multi-GPU distributed training scheme similar to~\cite{inceptionv4,xception}. The initial learning rate for this experiment was $0.1$ and was multiplied by 0.94 every 2 epochs. The epoch size is the same as the training set size with batch size $256$, momentum value $0.9$, and weight decay $0.0001$. The training data were augmented by random crop and horizontal flips. For proposed network learning, the weights corresponding to the concept outputs and the concept interconnections were learned for first 2 epochs before optimizing those for the category variables. We did not use label smoothing for training the Inception v4 model. The training was continued until the CNN achieved the same or close accuracy for category classification as reported in the original paper/repository.

\subsection{Analysis of MD-RN and MD-IC on ImageNet}\label{A:ANALYSIS}
\begin{figure}[h]
\vspace{-0.0in}
\begin{center}
\subfigure[\scriptsize $Acc_{\text{CAT}}$ distribution]{\includegraphics[width=0.49
\columnwidth,height=0.48\columnwidth]{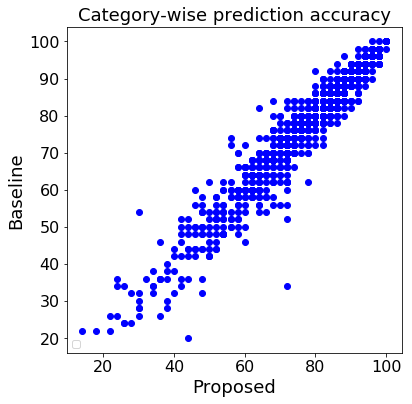}\label{F:ANALYSIS_DIST}}
\subfigure[\scriptsize $Acc_{\text{CAT}}$ progress]{\includegraphics[width=0.49
\columnwidth,height=0.48\columnwidth]{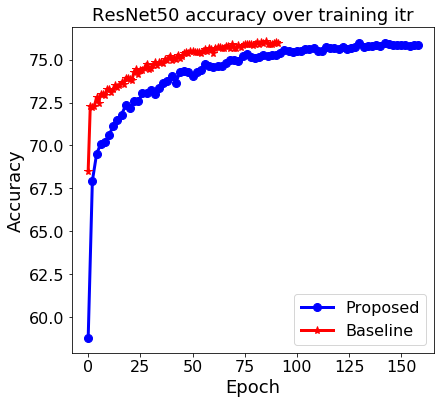}\label{F:ANALYSIS_PROG}}
\end{center}
\vspace{-0.5cm}
\caption{\small \subref{F:ANALYSIS_DIST} : Category-wise classification accuracy of the proposed method vs the baseline architecture (w/ ResNet50). \subref{F:ANALYSIS_PROG}: Progession of validation accuracy of the proposed CNN (blue) and baseline (red).}
\label{F:ANALYSIS}
\vspace{-0.1in}
\end{figure}

Since different category classes are predicted at different layers of the proposed dense structure, it is rational to verify whether or not the category classification capability is impaired by the depth of layer. To test this, we have plotted in Figure~\ref{F:ANALYSIS_DIST}  the category prediction accuracies for each of the $N=1000$ classes of the baseline against those of the proposed CNN with ResNet50 backbone. The plot implies no clear effect of the prediction depths on the classification performances on different categories as they remain same or very close to those of the original architecture.

\begin{figure*}[t]
\vspace{-0.0in}
\begin{center}
\includegraphics[width=1.95\columnwidth,height=0.42\columnwidth]{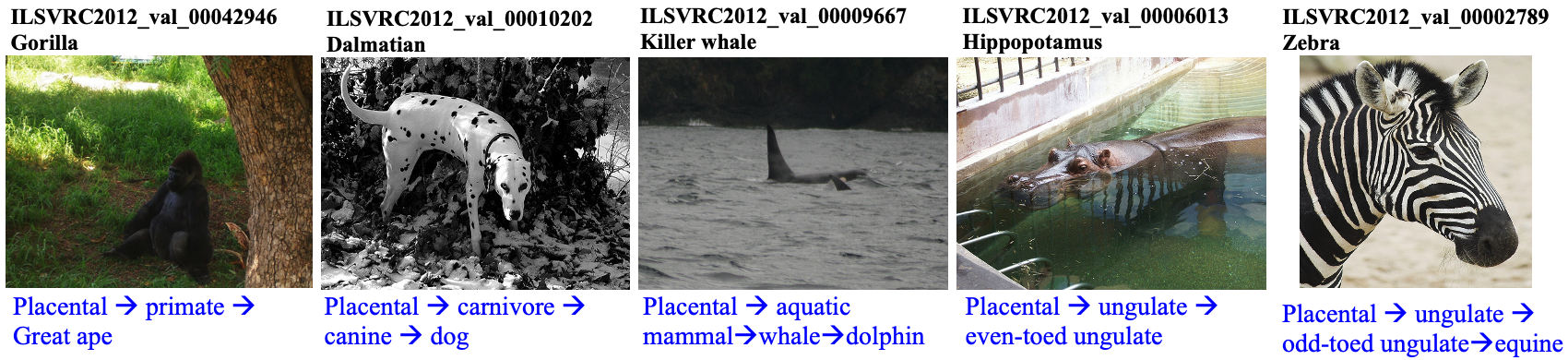}
\end{center}
\vspace{-0.5cm}
\caption{\small Qualitative performance of MD-RN trained from AwA2 datasets on images from ImageNet 12 validation set. The image ids and MD-RN category predictions are listed on top of each image whereas the concept predictions are displayed below.}
\label{F:RESULT_AWA_QUAL}
\vspace{-0.2cm}
\end{figure*}

Due to the increased number of dense layers and additional sigmoid activations, it is perhaps natural to expect the proposed architecture to require more iteration to converge. As Figure~\ref{F:ANALYSIS_PROG} demonstrates, our model indeed takes more epochs(x-axis) to attain a category classification performance (y axis) similar to that of baseline built upon ResNet50.

\subsection{Ablation studies}\label{A:ABLATION}
We have conducted ablation studies with model MD-RN.  In our first ablation study, we model the dense layers based on the original uncompressed hierarchy instead of the condensed hierarchy. Adopting the original ontology in ResNet-50 increased the number of z variables to 806, depth of dense layers to 17 and model size to 43M (compare with 40, 7, 30M resp. of compressed). Our attempts with different hyperparameters ($\lambda$, lr etc) achieved at most \{$Acc_{\text{CAT}}, Acc_{\text{CON}}, Acc_{\text{COMB}}$, mhP, mhR\}=\{44.02,1.12, 1.03, 60.05, 47.6\} at 30 epoch (compare $Acc_{\text{CAT}}$ with ours in Fig~\ref{F:ANALYSIS_PROG}). The training is perhaps smothered by 1) increased depth with many redundant single parent-child connections, and 2) imbalance in descendent distribution where smaller concept classes are under-represented.

The second analysis is performed with varying the values of the balancing term $\lambda$ in the combined loss function. As can be inferred from the network architecture for multiple dense connections and the loss functions, with $\lambda=0$ the category-wise loss drive concept $z$ variables of only the ancestor concepts to 1. However, with $\lambda =0$, the optimization will not force $z$ variables of the other non-ancestor concepts to $0$. Consequently, concept and combined accuracy of a model trained with $\lambda=0$ will be very low. 

As Table~\ref{T:ABLATION} shows, we observe a very low $Acc_{\text{CON}}, Acc_{\text{COMB}}$ for MD-RN-$\lambda0$ while achieving close $Acc_{\text{CAT}}$ to the published results. It is also apparent that a small $\lambda$ (= 2) favors improving  $Acc_{\text{CAT}}$ while a relatively large $\lambda$ (=8) favors improving $Acc_{\text{CON}}$ compared to those of the MD-RN model reported in the main submission. This is expected behavior with the loss function we are optimizing.
\begin{table} 
\vspace{-0.1cm}
\caption{\small Performance of MD-RN with different $\lambda$ values. All accuracies are computed for the single crop top-1 setting.}
\vspace{-0.2cm}
\begin{center}
\small
\begin{tabular}{|c|c|c|c|c|c|}
\hline
$\lambda$ & $Acc_{\text{CAT}}$ & $Acc_{\text{CON}}$ &$Acc_{\text{COMB}}$& mhP & mhR\\
\hline
0& 76.04 & 1.86 & 1.57 & 33.35 & 45.41\\
2& 76.04 & 78.57 & 66.66 &91.78 & 93.81\\
8& 75.42 & 81.74 & 70.08 & 94.13 & 94.08\\ 
\hline
\end{tabular}
\end{center}
\vspace{-0.4cm}
\label{T:ABLATION}
\end{table}

\subsection{Robustness of MD-RN trained on AwA2}\label{A:AWA}
In order to test the robustness of the proposed MD-RN network trained on AwA2 dataset, we evaluate its performance on ImageNet 12 images with overlapping categories. There are 16 AwA2 categories that are also present in ImageNet 12 dataset. We applied our MD-RN trained on AwA2 on 800 ImageNet 12 validation images and achieved category, concept and combined accuracies $\{ Acc_{\text{CAT}}, Acc_{\text{CON}}, Acc_{\text{COMB}} \} = \{ 89.87, 89.37, 85.75\}$. For comparison, we also applied the flat baseline BL-RN on these images and achieved $\{ Acc_{\text{CAT}}, Acc_{\text{CON}}, Acc_{\text{COMB}} \} = \{ 88.87, 78.12, 74.12\}$.  This experiment shows that the proposed dense layers in MD-RN learned on AwA2 can generalize better  than the single layer in BL-RN. Note that, the convolutional feature layers of both MD-RN and BL-RN have not changed because only dense layers of these two networks were trained. Figure~\ref{F:RESULT_AWA_QUAL} shows a few examples of output from the proposed network on ImageNet 12 validation images.

\subsection{Other Relevant Works}\label{A:RELWORKS}
Guo et al.~\cite{Guo2017CNNRNNAL} attempted to classify the coarse labels or the conceptual superclasses of categories by augmenting an RNN to CNN output. In addition to increased complexity imposed by the RNN, it is not clear how the hierarchy among labels was generated and how the hierarchy would scale up with increasing number of categories.  
Liang et al.~\cite{liang2019personalized} employed graph based reinforcement learning to learn features of a set of modules, each of which corresponds to a concept. Once the network search is completed, the activated module outputs are passed through prediction  layers for generating final output. The algorithm demonstrated promising performance for classifying scene contexts for semantic segmentation.

Commercial vision solutions such as Amazon Rekognition (https://aws.amazon.com/blogs/aws/amazon-rekognition-image-detection-and-recognition-powered-by-deep-learning/) or Google Vision ({https://cloud.google.com/vision}) seem to produce outputs that resemble concept classes. Due to the proprietary nature of these solutions, it is not possible for us to  confirm  whether or not they indeed predict concept classes and their relationships. However, based on the description provided in ({https://aws.amazon.com/blogs/aws/amazon-rekognition-image-detection-and-recognition-powered-by-deep-learning/}) and scrutiny of results, we speculate that Amazon rekognition does not use any hierarchy and probably uses a look-up from hard-coded (and undisclosed) relations strategy.

\end{document}